\documentclass[11pt]{article}

\usepackage[final]{acl}
\usepackage{dblfloatfix} 
\usepackage{times}
\usepackage{latexsym}
\usepackage{amsmath}
\usepackage[T1]{fontenc}
\usepackage[utf8]{inputenc}

\usepackage{microtype}
\usepackage{color}
\usepackage{inconsolata}

\usepackage{graphicx}
\usepackage{booktabs}
\usepackage{multirow}
\usepackage{subcaption} 
\usepackage[most]{tcolorbox}
\tcbuselibrary{listings,breakable, skins}

\usepackage[ruled,vlined]{algorithm2e}
\usepackage{xcolor}

\title{AgentCollab: A Self-Evaluation-Driven Collaboration Paradigm for Efficient LLM Agents}
\author{
\textbf{Wenbo Gao\textsuperscript{1,2}\thanks{Work done during internship at Huawei.}},
\textbf{Renxi Liu\textsuperscript{2}},
\textbf{Xian Wang\textsuperscript{2}},
\textbf{Fang Guo\textsuperscript{2}},
\textbf{Shuai Yang\textsuperscript{2}},
\textbf{Xi Chen\textsuperscript{2}},\\
\textbf{Hui-Ling Zhen\textsuperscript{2}},
\textbf{Hanting Chen\textsuperscript{2}},
\textbf{Weizhe Lin\textsuperscript{2}\thanks{Corresponding author.}},
\textbf{Xiaosong Li\textsuperscript{2}},
\textbf{Yaoyuan Wang\textsuperscript{2}}
\\
\textsuperscript{1}The Hong Kong Polytechnic University,
\textsuperscript{2}Huawei\\
\texttt{wenbo.gao@connect.polyu.hk}~~~~\texttt{linweizhe1@huawei.com}
}
\usepackage{amsmath} 
\usepackage{amssymb}
\usepackage{cleveref}
\usepackage{svg}
\begin{document}
\maketitle

\begin{abstract}
% Autonomous agents powered by large language models (LLMs) perform complex tasks through long-horizon reasoning and tool interaction, where a fundamental trade-off arises between execution efficiency and reasoning robustness. Small models enable fast inference but struggle in difficult reasoning segments, while large models provide stronger capabilities at substantially higher computational cost. We present \textit{AgentCollab}, a collaborative inference framework that dynamically coordinates language models of different sizes during agent execution. Instead of relying on external routing modules, the framework uses the model’s own self-reflection signal to determine whether the current reasoning step makes meaningful progress and escalates control to a larger model only when necessary. To further stabilize long-horizon execution, we introduce a difficulty-aware cumulative escalation strategy that allocates the large-model intervention budget based on recent failure signals. Experiments on diverse multi-step agent benchmarks show that AgentCollab consistently improves the accuracy–efficiency Pareto frontier of LLM agents.
Autonomous agents powered by large language models (LLMs) perform complex tasks through long-horizon reasoning and tool interaction, where a fundamental trade-off arises between execution efficiency and reasoning robustness. Models at different capability--cost levels offer complementary advantages: lower-cost models enable fast execution but may struggle on difficult reasoning segments, while stronger models provide more robust reasoning at higher computational cost. We present \textit{AgentCollab}, a self-driven collaborative inference framework that dynamically coordinates models with different reasoning capacities during agent execution. Instead of relying on external routing modules, the framework uses the agent’s own self-reflection signal to determine whether the current reasoning trajectory is making meaningful progress, and escalates control to a stronger reasoning tier only when necessary. To further stabilize long-horizon execution, we introduce a difficulty-aware cumulative escalation strategy that allocates additional reasoning budget based on recent failure signals. In our experiments, we instantiate this framework using a two-level small--large model setting. Experiments on diverse multi-step agent benchmarks show that AgentCollab consistently improves the accuracy--efficiency Pareto frontier of LLM agents. The code is available at https://github.com/SparkEngineAI/AgentFlashInfer.
\end{abstract}

\section{Introduction}

Autonomous agents powered by large language models (LLMs)~\cite{zhang2025agent, li2024survey} are increasingly deployed to tackle complex tasks such as web-based information seeking~\cite{zheng2025deepresearcher,qiao2025webresearcher,li2025websailor}, advanced mathematical reasoning~\cite{kapoor2026trim, yan2025mathagent}, and long-form writing~\cite{xia2025storywriter}. 
Unlike traditional single-turn question answering systems, these agents formulate plans, interact with external tools (e.g., web searching, document reading, and text composition), and iteratively refine their reasoning based on observed outcomes~\cite{shi2025pangu}. 
This iterative Think--Act--Observe loop enables agents to solve complex problems but also introduces substantial latency. 
An early mistake can propagate through subsequent steps and significantly degrade the final outcome~\cite{bopo,kapoor2026trim}. 
Allocating stronger reasoning capacity can mitigate such errors and improve reasoning quality. However, each reasoning step then becomes more expensive, and the accumulated latency grows rapidly in multi-turn scenarios. 
A fundamental dilemma therefore emerges between execution speed and reasoning quality.

Existing acceleration techniques mainly focus on improving efficiency for single-turn inference~\cite{xia2024speculative,egashira2024quantizatio} or short-horizon orchestration~\cite{ong2025routellm,chen2024frugalgpt,wang2025mixllm}. 
In contrast, the efficiency objective arises at a different level in autonomous agents due to long-horizon reasoning and interaction. 
Performance must be evaluated over the entire reasoning trajectory rather than individual inference steps. 
When an early mistake requires several additional cycles to correct, the time saved in a single step quickly becomes negligible compared with the extra turns spent on recovery. 
Correctness at a few critical steps often dominates overall efficiency, suggesting that optimization strategies should operate at the trajectory level instead of uniformly accelerating every step.

Recent work has begun to explore model routing as a way to balance capability and efficiency. 
Early studies mainly focus on \textit{single-turn} model selection, where a router chooses between cheaper and more capable models for a given query. 
Systems such as RouteLLM by~\citet{ong2025routellm} and FrugalGPT by~\citet{chen2024frugalgpt} employ lightweight routers to evaluate outputs from lower-cost models and invoke stronger models when necessary. 
These approaches primarily optimize monetary or token-level cost at the single-response level, which limits their relevance to long-horizon agents, where the central objective is end-to-end inference speed rather than per-query cost alone. 

Later work by~\citet{bopo} and~\citet{kapoor2026trim} extends routing to \textit{multi-step} agents by learning policies that select a model at each step of the reasoning process. 
To reduce inference cost, \citet{bopo} proposes allocating stronger models to steps predicted to be difficult but solvable. 
To improve reasoning quality, \citet{kapoor2026trim} leverages progress reward models to assess intermediate reasoning steps. 
While these methods move beyond single-turn routing, their main focus remains cost reduction or quality improvement. 
By contrast, our goal is to improve end-to-end reasoning speed in long-horizon agent execution. 
Under this objective, routing introduces additional challenges: routing decisions must be made repeatedly during execution, extra supervision is required to learn the routing policy, and frequent switching may interrupt computational reuse such as prefill states. 
The resulting overhead can offset the latency gains of lower-cost models, particularly in long reasoning trajectories. 
More fundamentally, for long-horizon agents the key challenge is not merely \textit{which model to call at a step}, but how control over the reasoning trajectory should be allocated and transferred across different levels of reasoning capacity.

These observations suggest a different perspective on efficiency. Instead of introducing external routing models or learned policies, the agent itself can already reflect on its reasoning process. That capability can serve as a natural signal for coordinating different levels of reasoning capacity. From this perspective, model collaboration can be driven internally by the reasoning process rather than controlled by an external router.

Following this intuition, we introduce a \textbf{self-driven collaboration paradigm} for multi-turn agents. 
In this paradigm, the reasoning process itself determines when stronger reasoning capacity is required. 
A lower-cost model tier performs most reasoning steps and continuously evaluates whether the current trajectory is making meaningful progress. 
When stagnation is detected, control is temporarily escalated to a stronger reasoning tier that resolves the difficult segment of the trajectory. 
Once progress resumes, control can return to a more efficient tier and the interaction loop continues. 
This process requires no auxiliary routing model and no additional policy training, as the decision signal is generated by the model’s own self-reflection.

Such collaboration naturally extends beyond a fixed pair of models. 
The same principle can coordinate a hierarchy of models with different capability--cost trade-offs, allowing reasoning capacity to be gradually increased only when necessary. 
This forms a scalable paradigm for allocating reasoning resources across long trajectories, where inexpensive tiers handle routine steps while stronger tiers intervene at critical moments.

Based on this paradigm, we further introduce a cumulative escalation strategy that adjusts the duration of higher-capacity intervention according to historical stagnation signals. 
Repeated failures lead to longer takeovers, enabling deeper correction on difficult trajectory segments while reducing oscillatory switching.

Our contributions are threefold:
\begin{itemize}
% \vspace{-5pt}
\item We introduce a \textbf{self-driven collaboration paradigm} for multi-turn LLM agents, in which the agent’s own trajectory-level reflection signals determine how control is allocated across models with different capability--cost trade-offs, without external routers or learned routing policies.
% \vspace{-5pt}\vspace{-5pt}
\item We propose a \textbf{difficulty-aware cumulative budget allocation strategy} that regulates higher-capacity intervention budget based on consecutive stagnation signals, enabling progressively stronger intervention under persistent reasoning difficulty.
% \vspace{-5pt}
\item We conduct experiments on diverse multi-turn agent benchmarks including web-based research, mathematical reasoning, and long-form writing. 
The results demonstrate a strong latency--quality trade-off, consistently improving reasoning efficiency while maintaining high performance.
\end{itemize}
\section{AgentCollab}
\label{sec:method}
% collab->cumulative escalation 震荡的问题large model prefill
This section introduces \textit{AgentCollab}, a collaborative efficient inference paradigm that dynamically allocates computational resources among language models of different sizes during multi-turn agent execution. The framework addresses a fundamental limitation of fixed-model agent systems. Small models offer fast inference but often struggle in complex reasoning segments, while large models provide stronger reasoning but incur substantially higher single-step latency when used throughout the entire trajectory. Relying on a single model therefore leads to an inherent trade-off between efficiency and robustness in long-horizon workflows.

\begin{figure*}[ht]
    \centering
  \includegraphics[width=0.95\linewidth]{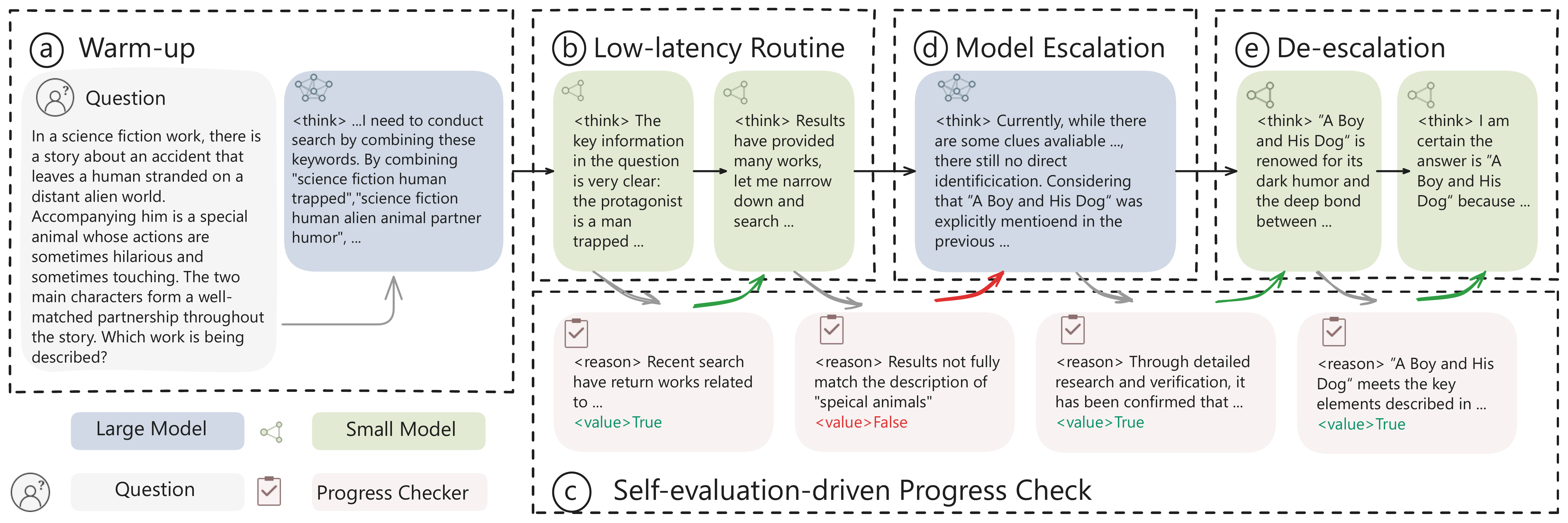}
  \caption{\textbf{Overview of the proposed AgentCollab framework.} (a) The agent system first invokes the large model to warm up the interaction and plan the overall strategy. (b) After initialization, the small model performs low-latency routine reasoning steps and (c) conducts self-evaluation on progress checks to determine whether escalation is required. (d) When stagnation is detected, control is temporarily transferred to a larger model within a predefined budget to resolve difficult reasoning segments. (e) Once the critical step is completed, control returns to the small model to continue the interaction.}
  \label{fig:overview}
  % \vspace{-10pt}
\end{figure*}

As illustrated in~\Cref{fig:overview}, AgentCollab introduces a model escalation mechanism that enables adaptive collaboration between small and large models during inference. The small model serves as the default controller for routine reasoning steps, while the large model is invoked only when the trajectory encounters difficult segments. 
%switch的instruction follow的准确率
%ddv2 multi agent best-N 采样弥补决策
%ddv2 
Building upon this mechanism, we further develop a \textit{difficulty-aware budget allocation strategy} that regulates the intervention strength of the large model. Instead of assigning a fixed intervention budget, the system adjusts the duration of large-model takeover based on recent failure signals along the trajectory. As shown in~\Cref{fig:allocation}, repeated stagnation progressively increases the escalation budget, enabling stronger corrective reasoning when the agent repeatedly fails to make progress.

% \subsection{Model Escalation}

% Autonomous agent systems execute multi-step reasoning procedures that involve planning, reflection, and tool interaction. Certain intermediate steps may become bottlenecks. The agent can stagnate at these stages and propagate errors to subsequent decisions. Uniform deployment of a large model mitigates this risk but increases latency. Uniform reliance on a small model improves speed but weakens accuracy. 

% To address this issue, we propose a model escalation strategy that coordinates heterogeneous language models through internal reflection. The small model acts as the default executor. After each reasoning step, it evaluates whether meaningful progress has been achieved toward the task objective. This self-assessment operates on the current reasoning state rather than external supervision. Limited progress indicates that the present capacity is insufficient.

% Upon detecting stagnation, control is transferred to the large model. The large model intervenes for several steps to resolve the challenging subtask and refine intermediate reasoning states. Then control is returned to the small model, which resumes inference from the improved state. This process concentrates high-capacity reasoning on difficult segments while preserving overall efficiency.

% Noted that, the first step is acted by the large model as the overall planning is commonly conducted at begining and it affect the overall accurcy severely. 
\subsection{Model Escalation}

Multi-step agent reasoning consists of sequential ``Think`` operations interleaved with tool interaction, during which certain trajectory segments may become locally trapped and lead to oscillation or redundant exploration. Persistent use of the large model $M_L$ reduces this risk at the expense of substantial computational cost, whereas exclusive reliance on the small model $M_S$ improves efficiency but weakens robustness in difficult regions.

To balance these effects, we design a \textbf{self-evaluation-driven escalation mechanism} that dynamically coordinates LLMs with different capabilities. As the early-stage reasoning strongly influences the subsequent trajectory and therefore benefits from higher-capacity inference~\citep{bopo,openpangu_deepdiver_v2}, the session begins with a short warm-up phase executed by strong model $M_L$. The global task framing, intermediate objective decomposition, and high-level strategy planning are performed in this phase. After initialization, control is transferred to small model $M_S$, which carries out the majority of later reasoning steps and tool calls in a low-latency routine.

To determine when to escalate, at each step $t$, the model produces not only a reasoning output but also a structured progress assessment, {which is generated in parallel with the normal reasoning output.} This reflection takes the form of a canonical block containing a \textit{rationale} field and a binary indicator \textit{value}. The rationale summarizes whether the latest reasoning steps advance the trajectory toward the final objective, while the indicator encodes this judgment as a binary variable $\mathrm{value}_t \in \{\texttt{TRUE}, \texttt{FALSE}\}$, where $\mathrm{value}_t = \texttt{TRUE}$ corresponds to meaningful progress and $\mathrm{value}_t = \texttt{FALSE}$ corresponds to stagnation. The signal is generated by the model itself based solely on the current reasoning context, and does not rely on any training process or auxiliary routing model.

When $\mathrm{value}_t = \texttt{TRUE}$, the controller maintains execution under the small model, allowing well-conditioned subproblems to be solved with minimal cost. When $\mathrm{value}_t = \texttt{FALSE}$, the trajectory is interpreted as entering a difficult region under the current capacity. The large model then takes over the control and executes for a bounded number of reasoning steps, during which it re-examines the accumulated context and restructures intermediate states if necessary. When the predefined intervention budget is exhausted or progress is made, execution subsequently returns to $M_S$. This bounded escalation–de-escalation process forms a closed feedback loop in which self-evaluation governs model routing. High-capacity computation is therefore concentrated on genuinely difficult segments while overall inference cost remains controlled.

\subsection{Difficulty-aware Budget Allocation Strategy}
\label{allocation}
The basic escalation mechanism determines when control should be transferred to the large model, yet it does not regulate the strength of intervention once escalation occurs. A small budget bound may lead to a weak intervention, failing to resolve persistent stagnation. However, an excessively large budget quota risks overusing escalation, increasing the latency instead.

\begin{figure*}[!htbp]
\centering
  \includegraphics[width=0.95\linewidth]{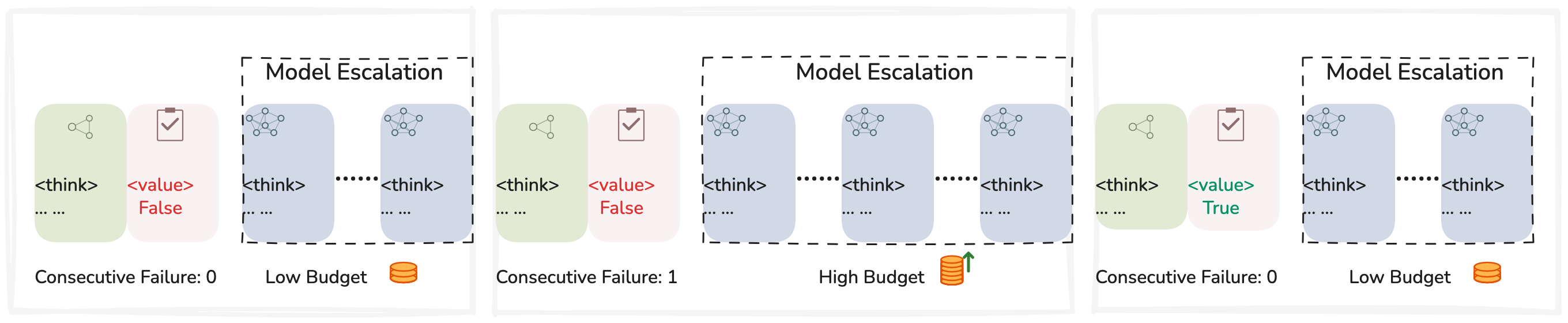}
  \caption{\textbf{Budget Allocation Strategy.} As consecutive failures accumulate, the framework temporarily increases the budget for the larger model, allowing stronger models to provide additional support.}
  \label{fig:allocation}
\end{figure*}

To address this limitation, a \textbf{difficulty-aware cumulative budget allocation strategy} that controls intervention depth at each escalation according to the recent self-evaluation history is proposed as shown in~\Cref{fig:allocation}. We first define the escalation level $l_t$ as the number of consecutive failure signals up to step $t$, which is recursively updated as
% \vspace{-6pt}
\begin{equation}
l_t =
\begin{cases}
l_{t-1} + 1, & \text{if } \mathrm{value}_t = \texttt{FALSE}, \\
0, & \text{if } \mathrm{value}_t = \texttt{TRUE}.
\end{cases}
\label{eq:level_update}
\end{equation}
The level therefore captures the accumulated difficulty reflected by trajectory-level self-evaluation.

Let $B^0$ denote the large-model intervention budget when $l=0$. 
{
The actual intervention horizon at escalation level $l$ is defined as a bounded growth function:
}
\begin{equation}
{
B^l = f(B^0,l) = \min\{B^{\max},\; B^0 + kl\},
}
\label{eq:sl_linear_bounded}
\end{equation}
{
where $k \ge 0$ controls the growth rate and $B^{\max}$ is the maximum intervention budget. 
When $k=0$, the strategy reduces to static model switching. 
When repeated stagnation increases $l$, the large model receives a longer but still bounded intervention budget, enabling stronger correction without unbounded latency growth.
}

{
Although $f(\cdot)$ may take other bounded forms, such as sigmoid or saturating schedules, we adopt the linear bounded formulation because it is simple, interpretable, and sufficient for controlling the intervention strength in our experiments.
}
\begin{table*}[t]
\small
\centering
\setlength{\tabcolsep}{4.2pt}
\begin{tabular}{clccccccccc}
\toprule
\multicolumn{1}{c}{} &
\multicolumn{1}{c}{} &
\multicolumn{3}{c}{\textbf{BrowseComp\_zh}} &
\multicolumn{3}{c}{\textbf{HLE-math}} &
\multicolumn{3}{c}{\textbf{WritingBench}} \\
\cmidrule(lr){3-5}
\cmidrule(lr){6-8}
\cmidrule(lr){9-11}
Agent &
Method &
Acc. (\%) &
\#Steps &
Speedup &
Acc. (\%) &
\#Steps &
Speedup &
Score &
\#Steps &
Speedup \\
\midrule
\multirow{9}{*}{DDV2}
& Small     & 18.3 & 21.75 & $\textbf{1.54}\times$ & 8.0  & 35.55 & $\textbf{3.38}\times$ & 4.4 & 17.25 & $\textbf{3.20}\times$ \\
& Large     & \textbf{34.6} & 17.29 & $1.00\times$ & \textbf{23.3} & 28.41 & $1.00\times$ & \textbf{5.1} & 9.10  & $1.00\times$ \\
& Random    & 22.3 & 20.56 & $1.12\times$ & 9.2  & 32.82 & $1.33\times$ & 4.2 & 13.10 & $1.29\times$ \\
& RouteLLM  & 28.2 & 22.40 & $1.09\times$ & 15.2 & 33.42 & $1.62\times$ & 4.7 & 10.27 & $2.21\times$ \\
& FrugalGPT & 27.2 & 20.20 & $1.29\times$ & 18.2 & 29.23 & $1.97\times$ & 4.5 & 12.22 & $1.94\times$ \\
& TRiM      & 29.6 & 21.38 & $1.18\times$ & 20.4 & 30.18 & $2.05\times$ & 4.6 & 12.86 & $1.71\times$ \\
& Miroflow  & 31.2 & 20.74 & $1.23\times$ & 18.7 & 31.64 & $1.84\times$ & 4.8 & 11.14 & $2.02\times$ \\
& WeDAS     & 30.8 & 19.96 & $1.28\times$ & 19.1 & 30.92 & $1.91\times$ & 4.8 & 10.92 & $2.08\times$ \\
& Ours      & \underline{33.9} & 19.82 & $\underline{1.36}\times$ & \underline{21.1} & 32.03 & $\underline{2.31}\times$ & \underline{5.0} & 9.59  & $\underline{2.43}\times$ \\
\midrule
\multirow{9}{*}{WebSailor}
& Small     & 14.2 & 10.38 & $\textbf{2.03}\times$ & 11.3 & 13.02 & $\textbf{2.48}\times$ & - & - & - \\
& Large     & \textbf{25.5} & 7.79  & $1.00\times$ & \textbf{14.0} & 7.09  & $1.00\times$ & - & - & - \\
& Random    & 14.9 & 10.48 & $1.20\times$ & 9.2  & 14.02 & $1.03\times$ & - & - & - \\
& RouteLLM  & 20.0 & 9.60  & $1.32\times$ & 11.4 & 9.34  & $1.21\times$ & - & - & - \\
& FrugalGPT & 19.7 & 10.50 & $1.28\times$ & 11.9 & 9.81  & $1.10\times$ & - & - & - \\
& TRiM      & 18.8 & 10.12 & $1.22\times$ & \underline{13.6} & 8.92  & $1.18\times$ & - & - & - \\
& Miroflow  & 21.4 & 9.18  & $1.35\times$ & 12.5 & 9.08  & $1.16\times$ & - & - & - \\
& WeDAS     & 21.0 & 9.06  & $1.39\times$ & 12.7 & 8.96  & $1.19\times$ & - & - & - \\
& Ours      & \underline{22.5} & 8.49  & $\underline{1.50}\times$ & 13.3 & 8.57  & $\underline{1.29}\times$ & - & - & - \\
\bottomrule
\end{tabular}
\caption{
\textbf{Main results on BrowseComp\_zh, HLE-math, and WritingBench.}
Speedup is measured relative to the large-model baseline in each agent framework.
AgentCollab consistently approaches large-model performance while preserving clear efficiency gains.
Results for WritingBench are not reported for WebSailor because it does not support long-form writing tasks. Best results are shown in \textbf{bold}, and second-best results are \underline{underlined}.
}
\label{tab:mainresults}
\vspace{-15pt}
\end{table*}
\section{Experiments and Discussion}
\subsection{Setup}
To evaluate the generality and effectiveness of the AgentCollab mechanism, we integrate it into two open-source agent frameworks, DDV2~\citep{openpangu_deepdiver_v2} and WebSailor~\citep{li2025websailor}, which provide pretrained models of different scales (7B/38B and 3B/7B/32B respectively). We evaluate the framework on diverse multi-turn tasks, including deep research (BrowseComp\_zh~\citep{zhou2025browsecomp}), challenging mathematical reasoning (HLE with math subjects~\citep{phan2025humanity}), and long-form generation (WritingBench~\citep{wu2025writingbench}). The three benchmarks contain 289, 866, and 1000 questions, respectively. {Appendix results on GAIA~\citep{mialon2024gaia} and XBench-DeepSearch~\citep{chen2025xbench} further cover additional deep-research benchmarks.} For fair comparison, all experiments use a maximum interaction budget of 40 iterations. The budget allocation strategy follows the linear formulation described in~\Cref{allocation}, with $B^0=2$ and $k=2$. All inference is conducted locally using vLLM-Ascend (v0.9.1) on Ascend 910B3 NPUs with four workers.

{For baselines, we compare AgentCollab with small-model-only and large-model-only agents, as well as several model-switching methods. \textit{Random} selects between the small and large models uniformly at random at each reasoning step. RouteLLM~\citep{ong2025routellm} and FrugalGPT~\citep{chen2024frugalgpt} were originally designed for cost-efficient single-turn routing, and we adapt them to the multi-turn setting by making routing decisions at each reasoning step. To cover multi-turn evaluation settings, we further include TRiM~\citep{kapoor2026trim}, Miroflow~\citep{su2026miroflow}, and WeDAS~\citep{yu2026WeDAS}. All switching baselines operate over the same small--large model pairs used by AgentCollab.}

Three metrics are reported: accuracy score, speedup, and the average number of reasoning iterations per question. Accuracy is evaluated by GPT-4o~\citep{achiam2023gpt}, while score is evaluated by the model provided by the benchmark~\citep{wu2025writingbench}. Speedup measures the end-to-end latency improvement relative to the large-model baseline and is defined as the ratio of the total latency of the large-model agent to that of the evaluated method. The iteration count reflects the reasoning cycles required to reach the final answer, with small models often requiring more rounds to recover from earlier mistakes.
\begin{figure}[!bp]
\centering
\includegraphics[width=0.95\columnwidth]{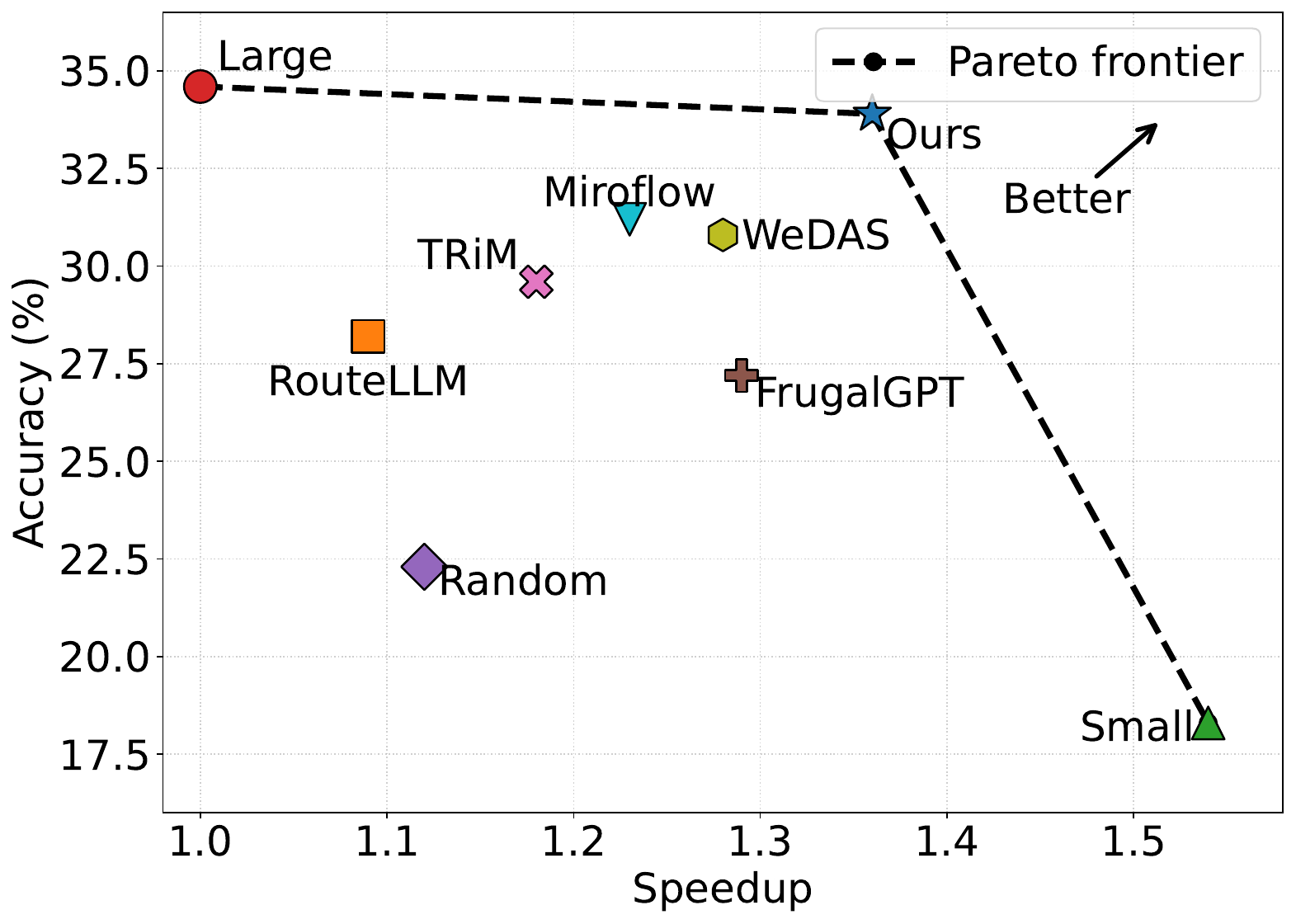}
\caption{Pareto frontier of DDV2 on BrowseComp\_zh.}
% \vspace{-15pt}
\label{fig:pareto}
\end{figure}

\begin{figure*}[!htbp]
    \centering
    \includegraphics[width=0.95\textwidth]{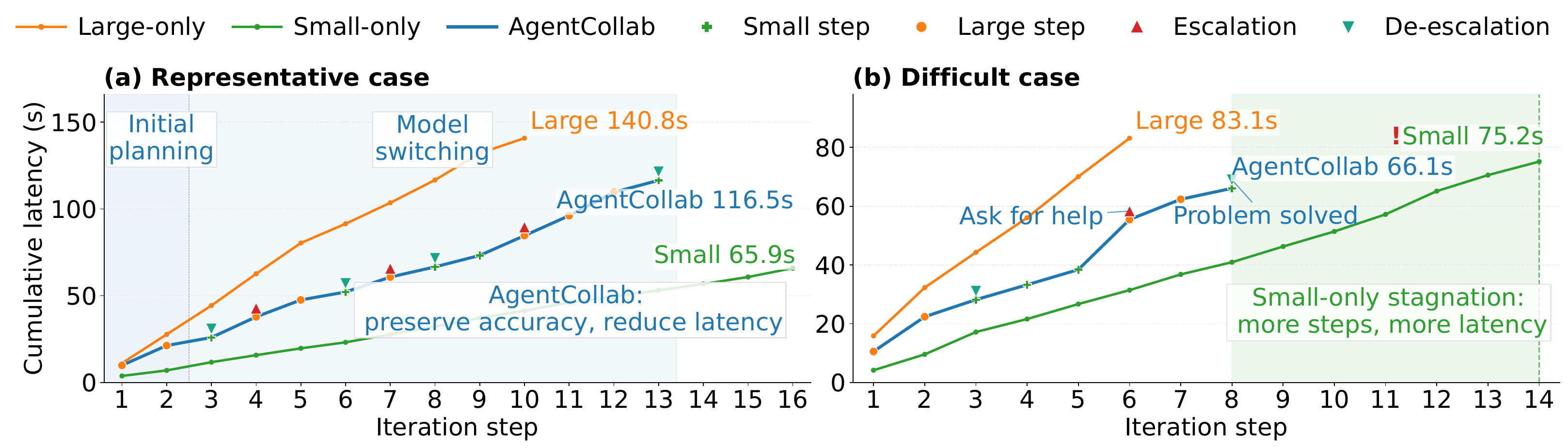}
\caption{
\textbf{Illustrative execution trajectories.}
Cumulative latency is shown across iteration steps for Large-only, Small-only, and AgentCollab. AgentCollab markers indicate the model used at each step, and triangles mark escalation or de-escalation events. In the representative case (a), AgentCollab reduces latency relative to Large-only execution by invoking the Large model only when necessary. In the difficult case (b), Large-only execution resolves the task within fewer steps, whereas Small-only execution becomes trapped in an unproductive reasoning trajectory and accumulates higher latency. AgentCollab escalates when needed and returns to the Small model after recovery.
}
    \label{fig:case_study}
    % \vspace{-15pt}
\end{figure*}
\subsection{Main Results}

\Cref{tab:mainresults} reports the results on three representative multi-turn agent benchmarks covering web research (BrowseComp\_zh), mathematical reasoning (HLE-math), and long-form generation (WritingBench). Overall, AgentCollab substantially improves small-model agents while preserving clear efficiency advantages over large-model agents.

As shown in~\Cref{tab:mainresults} and~\Cref{fig:pareto}, on BrowseComp\_zh, AgentCollab improves performance over the small-model agents for both DDV2 (18.3$\rightarrow$33.9) and WebSailor (14.2$\rightarrow$22.5), while remaining close to the large-model agents (34.6 and 25.5, respectively). Similar trends are observed on HLE-math, where collaboration substantially improves reasoning accuracy (e.g., $8.0\%\rightarrow21.1\%$ for DDV2) while still preserving clear speedup over the large-model baseline. On WritingBench, the collaborative system also improves generation quality (4.4$\rightarrow$5.0) and remains close to the large-model performance (5.1).

These results highlight a key challenge in multi-turn agents: although small models provide faster inference per step, they often require more interaction rounds to complete the task. AgentCollab mitigates this issue by invoking stronger models only when necessary, often leading to shorter reasoning trajectories while maintaining high efficiency.

Across all benchmarks, the collaborative approach achieves a favorable \emph{efficiency--quality frontier}: it substantially outperforms the small-model agents and approaches the performance of large-model agents, while preserving substantial efficiency gains over the large-model baseline. This demonstrates that the proposed self-reflection-driven escalation mechanism can effectively concentrate large-model computation on more difficult segments of the trajectory.

Compared with other model-switching baselines, AgentCollab achieves the strongest overall accuracy--efficiency trade-off. Random switching can achieve nontrivial speedup through frequent use of the small model, but the lack of trajectory awareness leads to weak reasoning performance. RouteLLM~\citep{ong2025routellm} and FrugalGPT~\citep{chen2024frugalgpt} improve over random switching by adapting single-turn routing mechanisms to step-level decisions, while TRiM~\citep{kapoor2026trim}, Miroflow~\citep{su2026miroflow}, and WeDAS~\citep{yu2026WeDAS} provide stronger multi-turn baselines. Nevertheless, these methods remain less effective than AgentCollab at allocating high-capacity computation along the full reasoning trajectory.
\subsection{Ablation Study}

To analyze the effectiveness of the proposed collaboration mechanism, we compare several configurations of small and large model allocation on the DDV2 agent system. The study examines the following aspects: the role assignment between planner and executor, the escalation strategy used in AgentCollab including the sensitivity of the budget growth parameter $k$.
\begin{table}[!htbp]
\centering
\small
\begin{tabular}{lcc}
\toprule
\textbf{Method} & \textbf{Acc. (\%)} & \textbf{Speedup} \\
\midrule
Large-model only & 34.6 & $1.00\times$ \\
Small-model only & 18.3 & $1.54\times$ \\
\midrule
\multicolumn{3}{l}{\textit{Role Assignment}} \\
\quad w/ Large Planner & 24.6 & $1.39\times$ \\
\quad w/ Large Executor & 27.3 & $1.24\times$ \\
\midrule
\multicolumn{3}{l}{\textit{Escalation Strategy}} \\
AgentCollab (Static)  & 32.5 & $1.32\times$ \\
AgentCollab (Dynamic) & 33.9 & $1.36\times$ \\
\bottomrule
\end{tabular}
\caption{\textbf{Ablation study on DDV2. }We examine the effect of planner/executor role assignment and escalation strategy. Accuracy and speedup are reported.}
\label{tab:ablation_main}
\vspace{-10pt}
\end{table}

\noindent\textbf{Impact of Planner and Executor Roles.} The DDV2 agent explicitly separates two roles: a \textit{planner} and an \textit{executor} (information seeker). The planner produces high-level reasoning plans, while the executor performs information retrieval and action execution. This structure allows us to examine how allocating the large model to different roles affects overall performance.

We enable the large model in these two roles separately. As the results shown in~\Cref{tab:ablation_main}, using the large model in the execution stage while keeping the planner as a small model yields higher accuracy than assigning the large model to the planner (27.3 vs.\ 24.6), although the speedup decreases slightly (from $1.39\times$ to $1.24\times$). This result suggests that the execution stage is the more error-sensitive component in multi-turn agents: it directly interacts with the environment, determines the quality of retrieved evidence, and more strongly influences the subsequent reasoning states. As a result, allocating stronger models to execution is more effective for improving the speed--quality frontier.

\begin{figure}[h]
\centering
\includegraphics[width=0.95\columnwidth]{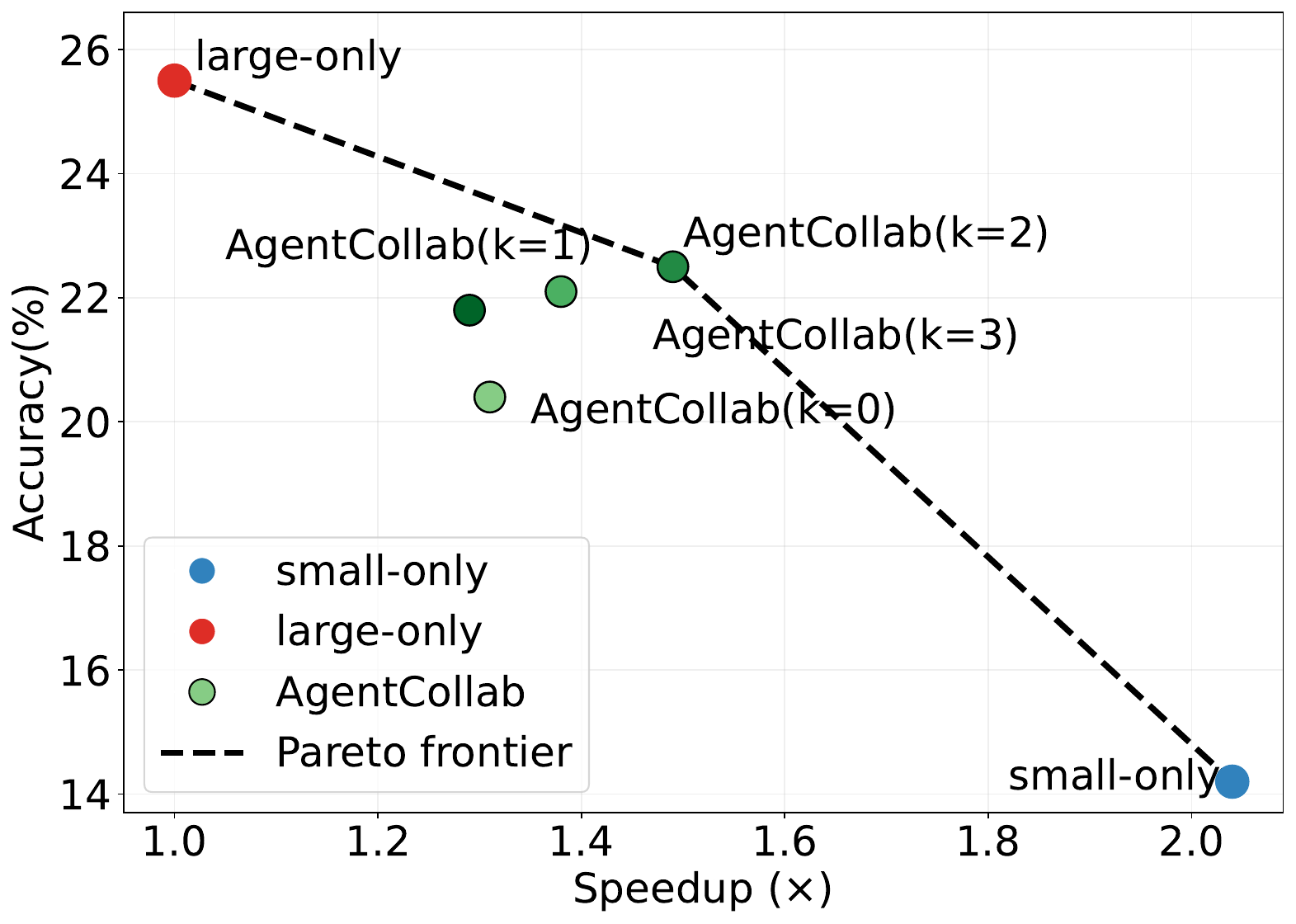}
\caption{Pareto frontier of AgentCollab within WebSailor on BrowseComp\_zh under different values of $k$.}
% \vspace{-15pt}
\label{fig:pareto_hyper}
\end{figure}

\noindent\textbf{Impact of Budget Allocation Strategy.}
{We further compare two escalation mechanisms in AgentCollab: a static allocation strategy and the proposed dynamic allocation strategy. Both configurations substantially outperform the small-model baseline, while the dynamic strategy achieves a better quality--efficiency trade-off than the static strategy. It improves accuracy from 32.5 to 33.9 and slightly increases speedup from $1.32\times$ to $1.36\times$, indicating that allocating larger intervention budgets after repeated stagnation is more effective than using a fixed escalation budget.

This improvement is also reflected in the switching behavior. A key motivation for the dynamic strategy is to reduce oscillatory switching, which can weaken large-model prefill caching by interrupting consecutive reasoning segments. We quantify this effect using the switching ratio defined in Equation~\ref{eq:switch_ratio} in the appendix. The dynamic strategy lowers the switching ratio from 49.64 to 45.07, suggesting fewer alternations between the small and large models. This supports our claim that difficulty-aware budget growth encourages more contiguous large-model intervention on difficult trajectory segments, thereby stabilizing execution and improving cache reuse.

To study budget allocation, we vary the hyperparameter $k$ from 0 to 3, where a larger $k$ yields more aggressive growth of the large-model intervention budget and $k=0$ reduces to the static strategy. A small budget may fail to resolve persistent stagnation, whereas an excessively large budget may overuse the large model and increase latency without proportional accuracy gains. The dynamic strategy balances these effects by increasing the intervention budget only after repeated stagnation, while keeping the total intervention bounded by the maximum budget. As shown in~\Cref{fig:pareto_hyper}, $k=2$ achieves the best quality--efficiency trade-off.}
% Smaller values provide insufficient correction for persistent stagnation, while larger values such as $k=3$ add computation with limited quality gain. 
This suggests that a moderate cumulative growth rate best balances corrective strength and efficiency.
% \vspace{-15pt}
\subsection{Analysis}
\noindent\textbf{Trajectory-Level Execution Analysis.}
{To better understand how AgentCollab behaves during multi-step execution, we visualize cumulative latency and model-escalation trajectories at the case level in Figure~\ref{fig:case_study}. 
In the representative case, AgentCollab reduces latency relative to Large-only execution by invoking the Large model only when necessary. 
The trajectory therefore grows more slowly than the Large-only baseline while still allowing stronger computation to intervene at selected steps.

The difficult case highlights why the fastest model at each step does not necessarily yield the most efficient trajectory. 
Small-only execution can become trapped in an unproductive reasoning path, requiring more iterations and accumulating higher total latency. 
AgentCollab avoids this inefficient trajectory by escalating when progress stalls and returning to the Small model after recovery. 
These examples suggest that end-to-end efficiency depends on both per-step cost and trajectory quality: a faster step is beneficial only when the reasoning process remains productive.}\\
% \vspace{-2pt}
\noindent\textbf{Impact of Agent Framework.}
We observe that AgentCollab performs less satisfactorily on WebSailor than on DDV2. One possible reason is the difference in instruction-following ability across the underlying models. WebSailor is built on an earlier Qwen2.5-based setup~\citep{qwen25}. An important observation is that WebSailor is more prone to false-positive progress judgments: even when the trajectory has not made meaningful progress, it is more likely to predict that progress has been made. Empirically, the false progress-check rate is around 3\% in DDV2~\citep{openpangu_deepdiver_v2}, compared with 7\% in WebSailor~\citep{li2025websailor}. In addition, DDV2 deploys a multi-process search agent, which provides more robust research results. {Weaker instruction following and less robust search make the self-evaluation signal less reliable and, in turn, weaken the effectiveness of model escalation. However, as shown in Table~\ref{tab:progress_check_recovery}, most progress-check misjudgments are corrected within the following one or two rounds, which limits their negative impact on the overall trajectory.}
\begin{table}[t]
\small
\centering
\begin{tabular}{lcc}
\toprule
\textbf{Agent} & \textbf{1-step Recovery} & \textbf{2-step Recovery} \\
\midrule
DDV2      & 96.8\% & 99.9\% \\
WebSailor & 95.2\% & 99.7\% \\
\bottomrule
\end{tabular}
\caption{
{\textbf{Recovery rate after false-positive progress judgments.}
Measured by manual inspection, most misjudgments are corrected within one step, and nearly all are corrected within two steps.}}
\vspace{-10pt}
\label{tab:progress_check_recovery}
\end{table}

{\noindent\textbf{Progress Check Frequency.} Although the progress signal is generated in parallel with the normal reasoning output rather than through an additional inference call, its overhead is worth examining. We therefore evaluate whether reducing the frequency of self-evaluation can improve efficiency. As shown in Table~\ref{tab:progress_check_frequency}, less frequent progress checks only slightly improve the speedup, but this gain is accompanied by more reasoning steps and substantially lower accuracy in both DDV2 and WebSailor. The modest speedup mainly comes from increased small-model usage, as delayed checking postpones large-model intervention and leaves more iteration steps to the small model. However, this also allows unproductive or incorrect trajectories to persist longer before correction, offsetting the small saving from less frequent checking. Therefore, reducing progress-check frequency does not provide a clear end-to-end benefit, and we use step-wise progress checking as the default setting.
\begin{table}[t]
\small
\centering
\setlength{\tabcolsep}{3pt} % 默认约6pt，可继续调小
\begin{tabular}{clccc}
\toprule
\textbf{Agent} &
\textbf{Frequency} &
\textbf{Acc. (\%)} &
\textbf{\#Steps} &
\textbf{Speedup} \\
\midrule
\multirow{3}{*}{DDV2}
& Every step    & \textbf{33.9} & \textbf{19.82} & 1.36$\times$ \\
& Every 3 steps & 32.7          & 20.46          & 1.40$\times$ \\
& Every 5 steps & 31.8          & 21.17          & \textbf{1.43}$\times$ \\
\midrule
\multirow{3}{*}{WebSailor}
& Every step    & \textbf{22.5} & \textbf{8.49} & 1.50$\times$ \\
& Every 3 steps & 20.7          & 8.93          & 1.55$\times$ \\
& Every 5 steps & 19.4          & 9.48          & \textbf{1.59}$\times$ \\
\bottomrule
\end{tabular}
\caption{
\textbf{Ablation on progress-check frequency on BrowseComp\_zh.}
Speedup is measured relative to the large-model baseline in each agent framework.
}
\vspace{-15pt}
\label{tab:progress_check_frequency}
\end{table}

\noindent\textbf{Heterogeneous Model Pairing.}
We further evaluate collaboration between models that differ not only in size but also in model family. As shown in Table~\ref{tab:heterogeneous_ddv2}, same-family pairings show a slight advantage, suggesting that shared training backgrounds may lead to more consistent reasoning styles, tool-use behaviors, and progress-check interpretations. Such cross-model alignment remains challenging in heterogeneous model collaboration~\citep{chen2026heter,hu2025efficient}. Nevertheless, the gap is modest: Pangu-7B + Qwen-32B achieves 31.9 accuracy with $1.41\times$ speedup, and Qwen-7B + Pangu-38B achieves 32.6 accuracy with $1.44\times$ speedup. These results suggest that AgentCollab is not limited to the same-family size scaling and can operate across models with different pretraining backgrounds.}
\begin{table}[t]
\small
\centering
\setlength{\tabcolsep}{2pt}
\begin{tabular}{lccc}
\toprule
\textbf{Model Pair} &
\textbf{Acc. (\%)} &
\textbf{\#Steps} &
\textbf{Speedup} \\
\midrule

\multicolumn{4}{l}{\textit{Baseline same-family pairing}} \\
Pangu-7B + Pangu-38B           & 33.9 & 19.82          & 1.36$\times$ \\
Qwen-7B + Qwen-32B             & 32.4          & 17.64 & 1.49$\times$ \\
\midrule

\multicolumn{4}{l}{\textit{Cross-family pairing}} \\
Pangu-7B + Qwen-32B            & 31.9          & 18.72          & 1.41$\times$ \\
Qwen-7B + Pangu-38B            & 32.6          & 18.05          & 1.44$\times$ \\

\bottomrule
\end{tabular}

\caption{
\textbf{Heterogeneous model-pair ablation under the DDV2 framework on BrowseComp\_zh.}
The first block reports same-family baseline pairings, while the second block evaluates cross-family collaborations. 
Cross-family pairings retain competitive accuracy and speedup, suggesting that AgentCollab remains feasible beyond same-family model collaboration.
}
\label{tab:heterogeneous_ddv2}
\vspace{-15pt}
\end{table}

\noindent\textbf{Failure analysis.} To better understand the remaining limitations of AgentCollab, we compare successful and failed cases in terms of reasoning steps, large-token ratio, and total latency. Here, the large-token ratio denotes the proportion of tokens generated by the large model and serves as a proxy for reliance on the large model. Figure~\ref{fig:failure-analysis} {in the appendix} shows that failed cases generally involve longer trajectories, greater reliance on the large model, and higher end-to-end latency.
\section{Conclusion}
% \vspace{-2pt}
This paper presents \textit{AgentCollab}, a collaborative inference framework that dynamically coordinates language models of different sizes for long-horizon agent reasoning. Instead of relying on external routing modules or fixed model allocation, the framework adopts a self-evaluation-driven escalation mechanism, allowing the model itself to determine whether the current reasoning trajectory is making meaningful progress and whether stronger intervention is needed. To improve multi-step execution, we incorporate a difficulty-aware budget allocation strategy that adaptively regulates large-model intervention under persistent stagnation.
Experimental results show that AgentCollab achieves a favorable trade-off between reasoning quality and execution latency, improving the Pareto frontier across diverse agent benchmarks. More broadly, the proposed framework points to a promising direction for efficient agent systems, where heterogeneous language models collaborate through internal signals rather than external orchestration.
\section*{Limitations}
The main efficiency evaluation is conducted with open-source LLMs deployed through local inference services, where latency, batching behavior, hardware allocation, and decoding configurations can be controlled consistently across methods. We also provide additional experiments with closed-source API models, suggesting that the proposed collaboration mechanism remains applicable in API-based settings. However, API latency is affected by server-side scheduling and network conditions, making it difficult to fairly compare end-to-end speedup with locally deployed models. We therefore report accuracy and reasoning steps for API-based experiments, while leaving a more controlled analysis of API latency and monetary cost to future work.

\bibliography{anthology}

\newpage
\appendix
\section{Additional Benchmark}
\label{sec:app_additional_benchmark}

To further examine whether the observed accuracy--efficiency trade-off generalizes beyond the main benchmarks, we report additional results on GAIA and XBench in Table~\ref{tab:mainresults_additional_deepsearch}. These benchmarks cover complementary multi-step reasoning and tool-use scenarios. The results show a consistent trend with the main experiments: AgentCollab improves over the small-model baseline while retaining clear efficiency advantages over the large-model baseline, suggesting that the proposed collaboration mechanism is not specific to a single benchmark or agent framework.

\begin{table*}[t]
\small
\centering
\setlength{\tabcolsep}{7pt}
\begin{tabular}{clcccccc}
\toprule
\multicolumn{1}{c}{} &
\multicolumn{1}{c}{} &
\multicolumn{3}{c}{\textbf{GAIA}} &
\multicolumn{3}{c}{\textbf{xbench-DeepSearch}} \\
\cmidrule(lr){3-5}
\cmidrule(lr){6-8}
Agent &
Method &
Acc. (\%) &
\#Steps &
Speedup &
Acc. (\%) &
\#Steps &
Speedup \\
\midrule
\multirow{9}{*}{DDV2}
& Small     & 35.9 & 24.18 & $\textbf{1.68}\times$ & 39.0 & 27.32 & $\textbf{1.82}\times$ \\
& Large     & \textbf{54.2} & 18.52 & $1.00\times$ & \textbf{53.0} & 20.74 & $1.00\times$ \\
& Random    & 38.6 & 23.04 & $1.19\times$ & 41.8 & 26.15 & $1.17\times$ \\
& RouteLLM  & 45.1 & 24.58 & $1.16\times$ & 45.6 & 27.40 & $1.14\times$ \\
& FrugalGPT & 44.3 & 22.36 & $1.33\times$ & 46.4 & 24.92 & $1.31\times$ \\
& TRiM      & 45.8 & 23.21 & $1.24\times$ & 47.2 & 25.84 & $1.23\times$ \\
& Miroflow  & 49.6 & 21.84 & $1.31\times$ & 49.8 & 24.31 & $1.34\times$ \\
& WeDAS     & 48.9 & 21.27 & $1.37\times$ & 49.1 & 23.76 & $1.40\times$ \\
& Ours      & \underline{52.3} & 21.03 & $\underline{1.45}\times$ & \underline{51.7} & 23.18 & $\underline{1.52}\times$ \\
\midrule
\multirow{9}{*}{WebSailor}
& Small     & 37.9 & 12.04 & $\textbf{2.16}\times$ & 34.3 & 13.47 & $\textbf{2.29}\times$ \\
& Large     & \textbf{53.2} & 8.31  & $1.00\times$ & \textbf{53.3} & 8.86  & $1.00\times$ \\
& Random    & 40.7 & 12.28 & $1.22\times$ & 37.5 & 13.86 & $1.15\times$ \\
& RouteLLM  & 45.6 & 10.76 & $1.34\times$ & 43.2 & 11.72 & $1.28\times$ \\
& FrugalGPT & 44.9 & 11.34 & $1.29\times$ & 42.6 & 12.36 & $1.21\times$ \\
& TRiM      & 44.1 & 11.06 & $1.25\times$ & 41.8 & 12.08 & $1.19\times$ \\
& Miroflow  & 48.7 & 10.18 & $1.38\times$ & 47.4 & 10.94 & $1.36\times$ \\
& WeDAS     & 48.2 & 9.92  & $1.43\times$ & 46.8 & 10.62 & $1.41\times$ \\
& Ours      & \underline{51.0} & 9.36  & $\underline{1.57}\times$ & \underline{50.6} & 10.03 & $\underline{1.55}\times$ \\
\bottomrule
\end{tabular}
\caption{
\textbf{Additional results on GAIA and xbench-DeepSearch.}
Speedup is measured relative to the large-model baseline in each agent framework.
AgentCollab maintains strong accuracy-efficiency trade-offs on additional deep-search benchmarks.
}
\label{tab:mainresults_additional_deepsearch}

\end{table*}

\section{Closed-source API}
\label{sec:app_api}

We also evaluate AgentCollab with closed-source API models, as reported in Table~\ref{tab:closed_source_ddv2_browsecomp}. This setting tests whether the proposed collaboration mechanism remains effective when the underlying models are accessed only through black-box inference APIs. The results show that AgentCollab can still provide a favorable balance between task performance and inference cost, indicating that the method is compatible with both open-source local deployment and practical API-based agent systems.

\begin{table}[t]
\small
\centering
\begin{tabular}{lcc}
\toprule
\textbf{Model Pair} &
\textbf{Acc. (\%)} &
\textbf{\#Steps} \\
\midrule
GPT-5-mini + GPT-5 & \textbf{32.8} & \textbf{20.35} \\
Gemini-3-Flash + Gemini-3-Pro & 31.6 & 20.91 \\
\bottomrule
\end{tabular}
\caption{
\textbf{Closed-source API model-pair results on BrowseComp\_zh under the DDV2 framework.}
We report accuracy and reasoning steps only, since API latency is affected by server-side scheduling and network conditions.
}
\label{tab:closed_source_ddv2_browsecomp}
\end{table}

\section{Definition of the Switching Ratio}
The switching ratio used to quantify oscillatory switching is defined as:
\begin{equation}
R_{\text{switch}}
=
\frac{\sum_{i=1}^{N}\sum_{t=2}^{T_i}\mathbb{I}(m_{i,t}\neq m_{i,t-1})}
{\sum_{i=1}^{N}(T_i-1)}.
\label{eq:switch_ratio}
\end{equation}
\noindent where \(N\) is the number of trajectories, \(T_i\) is the number of steps in trajectory \(i\), \(m_{i,t} \in \{M_S, M_L\}\) denotes the selected model at step \(t\), and \(\mathbb{I}(\cdot)\) is the indicator function.

\section{Case Study}
To further illustrate how AgentCollab corrects stalled reasoning trajectories through model escalation, we present a representative case in~\Cref{tab:case_study}, comparing small-model-only, large-model-only, and AgentCollab under the same task. The small-model-only setting becomes trapped in a difficult subproblem and gradually drifts toward an incorrect search direction, whereas the large-model-only setting is able to maintain broad and accurate exploration throughout the trajectory. AgentCollab initially encounters a similar difficulty, but detects the lack of meaningful progress through self-evaluation and temporarily transfers control to the large model. After the difficult segment is resolved, control returns to the small model, which resumes efficient reasoning on the corrected trajectory. This example shows that the proposed collaboration mechanism can recover from local stagnation while preserving much of the efficiency advantage of the small model.
\begin{table*}[htbp]
\centering
\small
\begin{tabular}{p{0.3\linewidth} p{0.3\linewidth} p{0.3\linewidth}}
\toprule
\textbf{Small Model} &
\textbf{AgentCollab} &
\textbf{Large Model} \\
\midrule

% ----------- Step 1 -----------
\textcolor{blue}{
Let me analyze this task: The task requires searching for a historically significant document jointly signed by multiple national representatives, with a focus on the English copy stored in a Chinese museum. I should begin with a web search.
}
&
\textcolor{blue}{
Let me analyze this task: I need to search for a multi-nation jointly signed historical document, stored in a Chinese museum, in English. Possible candidates include the UN Charter, Treaty of Versailles, etc. I need both Chinese and English keywords.
}
&
\textcolor{blue}{
Let me analyze this task: Find an English document jointly signed by multiple national representatives with major historical significance, and stored in a Chinese museum. I should focus on international treaties, UN documents, and museum collections in China.
}
\\

\multicolumn{3}{p{0.95\linewidth}}{
\textbf{Analysis:}
All three systems begin with high-quality reasoning from the large model. \textit{The Small Model mode also starts with large model planning to compare them in a more fair configuration.}
}
\\
\midrule
......
&
......
&
......
\\
\midrule

% ----------- Step 3 -----------
\textcolor{red}{
Now I have obtained detailed information on the Cairo Declaration… Let me search for other similar important documents.
}
&
\textcolor{red}{
I still have not found a multi-nation signed English document stored in a Chinese museum. Let me try searching for the “UN Conference Memorial Album” from the Museum of Overseas Chinese History.
}
\textcolor{orange}{
\textbf{[Progress check: no significant progress made. Escalate.]}
}
&
\textcolor{blue}{
I found important information: UN Charter drafting documents (Chongqing Library), “UN Conference Memorial Album” (Museum of Overseas Chinese History), and other multi-nation historical materials.
}
\\

\multicolumn{3}{p{0.95\linewidth}}{
\textbf{Analysis:}
The left system enters a local loop with limited search breadth; the AgentCollab system encounters stagnation and triggers model escalation; the right system continues high-quality knowledge expansion.
}
\\
\midrule
......
&
......
&
......
\\
\midrule
% ----------- Step 4 -----------
\textcolor{red}{
From the second round of search results, I did not find other documents that clearly meet the requirements… Let me search the UN Charter or other international treaties.
}
&
\textcolor{blue}{
Great! I found the “UN Conference Memorial Album”: stored in the Museum of Overseas Chinese History, involving 50 nations, containing bilingual UN Charter texts, fully meeting the task requirements.
}
\textcolor{orange}{
\textbf{[Restore to small model.]}
}
&
\textcolor{blue}{
Summary of findings: English UN Charter, UN Conference Memorial Album, “Records of Liberated Areas of China,” and other relevant documents.
}
\\

\multicolumn{3}{p{0.95\linewidth}}{
\textbf{Analysis:}
This is where collaborative reasoning shows its clear advantage: the left system fails to find the key document; the middle system, after escalation, finds the exact same crucial evidence as the large-model-only system.
}
\\
\midrule
......
&
......
&
......
\\
\midrule
% ----------- Step 5 -----------
\textcolor{red}{
Now I have enough information to complete the task: Cairo Declaration, UN Charter (museum collection unclear).
}
&
\textcolor{red}{
I found three classes of documents: the UN Conference Memorial Album; UN documents stored in the National Library of China; and international convention files stored through UN depository library systems (including English versions).
}
&
\textcolor{blue}{
Let me reorganize all findings: English UN Charter, UN Conference Memorial Album, Records of Liberated Areas of China, and UN Charter drafting documents — ready to finish the task.
}
\\

\multicolumn{3}{p{0.95\linewidth}}{
\textbf{Analysis:}
The right system maintains maximal coverage at high cost; the middle system reaches nearly the same breadth and accuracy through collaboration; the left system misses the core target and yields the lowest final quality.
}
\\

\bottomrule
\end{tabular}
\caption{\textbf{Comparison of key THINK steps across three reasoning modes}: small-model-only, AgentCollab, and large-model-only.}
\label{tab:case_study}
\end{table*} 
% \section{Example Appendix}
% \label{sec:appendix}
% \section{Faliure Analysis}
\begin{figure*}[ht]
  \centering
  \begin{subfigure}[h]{0.25\textwidth}
    \centering
    \includegraphics[width=\linewidth]{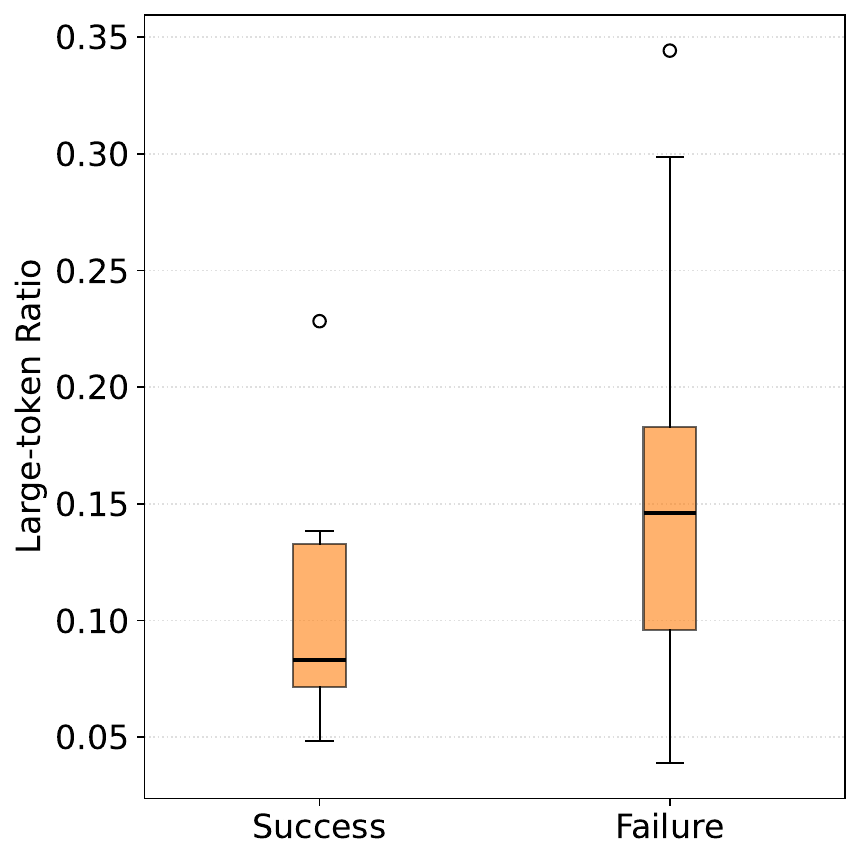}
    \caption{Large-model-token Ratio}
    \label{fig:failure-ltr}
  \end{subfigure}
  \hfill
  \begin{subfigure}[h]{0.25\textwidth}
    \centering
    \includegraphics[width=\linewidth]{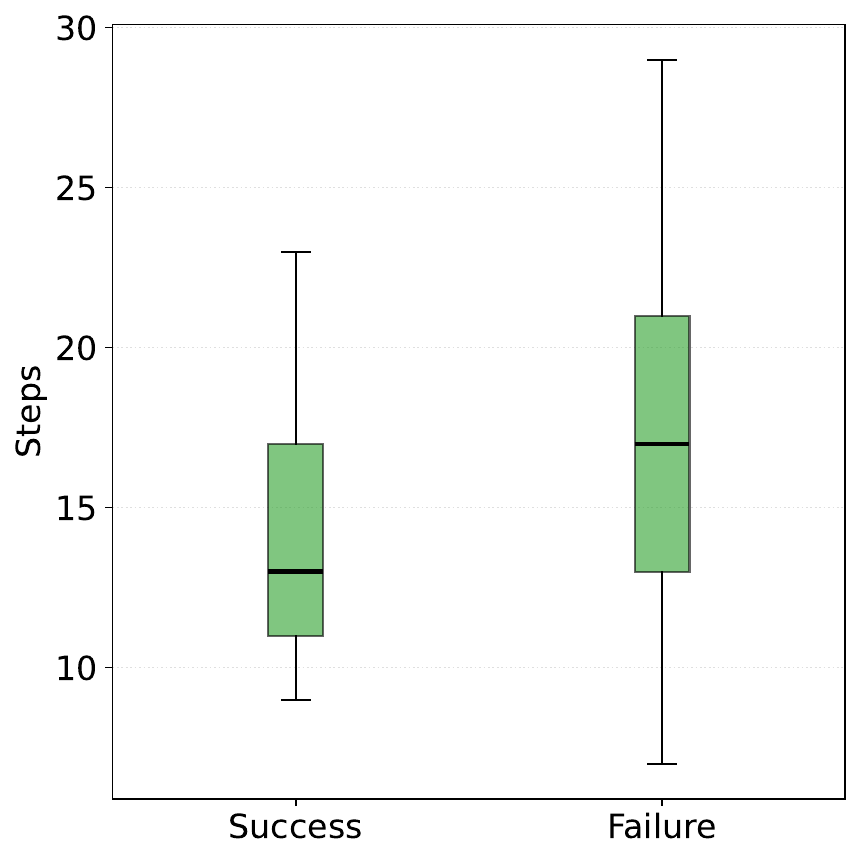}
    \caption{Number of Steps}
    \label{fig:failure-steps}
  \end{subfigure}
  \hfill
  \begin{subfigure}[h]{0.25\textwidth}
    \centering
    \includegraphics[width=\linewidth]{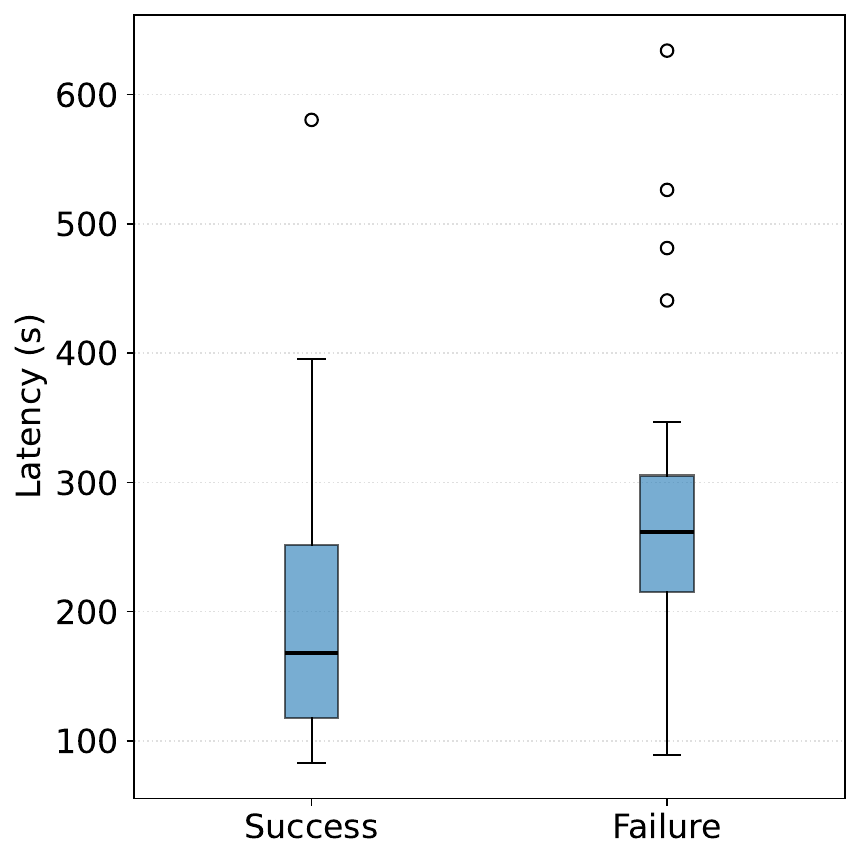}
    \caption{Latency}
    \label{fig:failure-latency}
  \end{subfigure}

 \caption{\textbf{Failure analysis.} Box plots comparing successful and failed cases in terms of (a) large-token ratio (the proportion of tokens generated by the Large model), (b) number of reasoning steps, and (c) latency.}
  %Failed cases tend to exhibit higher large-token ratios, more steps, and higher overall latency, suggesting that failures are associated with harder tasks that require increased use of the Large model and longer execution.
  \label{fig:failure-analysis}
  % \vspace{-15pt}
\end{figure*}

\begin{table}[t]
\centering
\small
\begin{tabular}{lccc}
\toprule
\textbf{Configuration} & \textbf{Acc. (\%)} & \textbf{\#Step} & \textbf{Speedup} \\
\midrule
3B                    & 9.7  & 9.44  & 1.37$\times$ \\
3B + 7B               & 12.5 & 11.40 & 1.08$\times$ \\
7B                    & 14.2 & 10.38 & -- \\
\midrule
3B + 32B              & 12.8 & 9.08  & 1.40$\times$ \\
7B + 32B              & 22.5 & 8.49  & 1.50$\times$ \\
32B                   & 25.5 & 7.79  & -- \\
\bottomrule
\end{tabular}
\caption{\textbf{WebSailor on BrowseComp\_zh with different model size configurations.}
Configurations such as 3B + 7B denote collaboration between two models of different sizes. Speedup is computed against 7B in the upper block and against 32B in the lower block.}
\label{tab:websailor_model_sizes}
% \vspace{-5pt}
\end{table}
\section{Difficulty-Split Analysis of Progress-Check Frequency}
\label{app:progress_check_difficulty}

To further examine how progress-check frequency affects tasks with different reasoning complexity, we split the BrowseComp\_zh instances into simple and hard subsets according to the number of reasoning steps required by the default step-wise progress-check setting. Specifically, tasks with fewer than 10 steps for DDV2 and fewer than 5 steps for WebSailor are treated as simple tasks, while the remaining tasks are treated as hard tasks. This analysis complements the main progress-check frequency ablation by showing whether the observed accuracy--efficiency trade-off differs across task difficulty levels.

\begin{table*}[!htbp]
\small
\centering
\begin{tabular}{clcccc}
\toprule
\textbf{Agent} &
\textbf{Task Type} &
\textbf{Acc. (3-step)} &
\textbf{Speedup} &
\textbf{Acc. (5-step)} &
\textbf{Speedup} \\
\midrule
\multirow{2}{*}{DDV2}
& Simple & {33.6 (-0.3)} & {1.37$\times$} & {33.3 (-0.6)} & {1.39$\times$} \\
& Hard   & {33.0 (-0.9)} & {1.42$\times$} & {32.4 (-1.5)} & {1.47$\times$} \\
\midrule
\multirow{2}{*}{WebSailor}
& Simple & {22.1 (-0.4)} & {1.52$\times$} & {21.8 (-0.7)} & {1.54$\times$} \\
& Hard   & {21.1 (-1.4)} & {1.58$\times$} & {20.1 (-2.4)} & {1.63$\times$} \\
\bottomrule
\end{tabular}
\caption{
{\textbf{Difficulty-split analysis of progress-check frequency on BrowseComp\_zh.}
Accuracy drops relative to step-wise progress checking are shown in parentheses. Simple tasks achieve limited speedup with minimal accuracy degradation, whereas hard tasks gain more efficiency but suffer larger performance drops.}}
\label{tab:progress_check_difficulty}
\end{table*}

As shown in Table~\ref{tab:progress_check_difficulty}, reducing the progress-check frequency has a smaller effect on simple tasks, where the reasoning horizon is short and delayed escalation is less likely to affect the final answer. By contrast, hard tasks obtain larger speedup because more reasoning steps are executed by the small model before escalation is triggered, but this also leads to larger accuracy degradation. These results are consistent with the main ablation in Table~\ref{tab:progress_check_frequency}: less frequent checking can slightly improve efficiency, but delayed intervention increases the risk that the small model remains on an unproductive trajectory for longer. We therefore use step-wise progress checking as the default setting.
\section{Difficulty-Aware Speedup Analysis} To better understand how the efficiency gain varies across task complexity, we analyze the per-instance speedup ratio with respect to task difficulty. 
Following a simple operational proxy, we use the execution latency of the large model as an estimate of task difficulty: instances that require longer large-model execution are treated as more difficult.

\begin{figure}[h]
  \centering

    \includegraphics[width=\columnwidth]{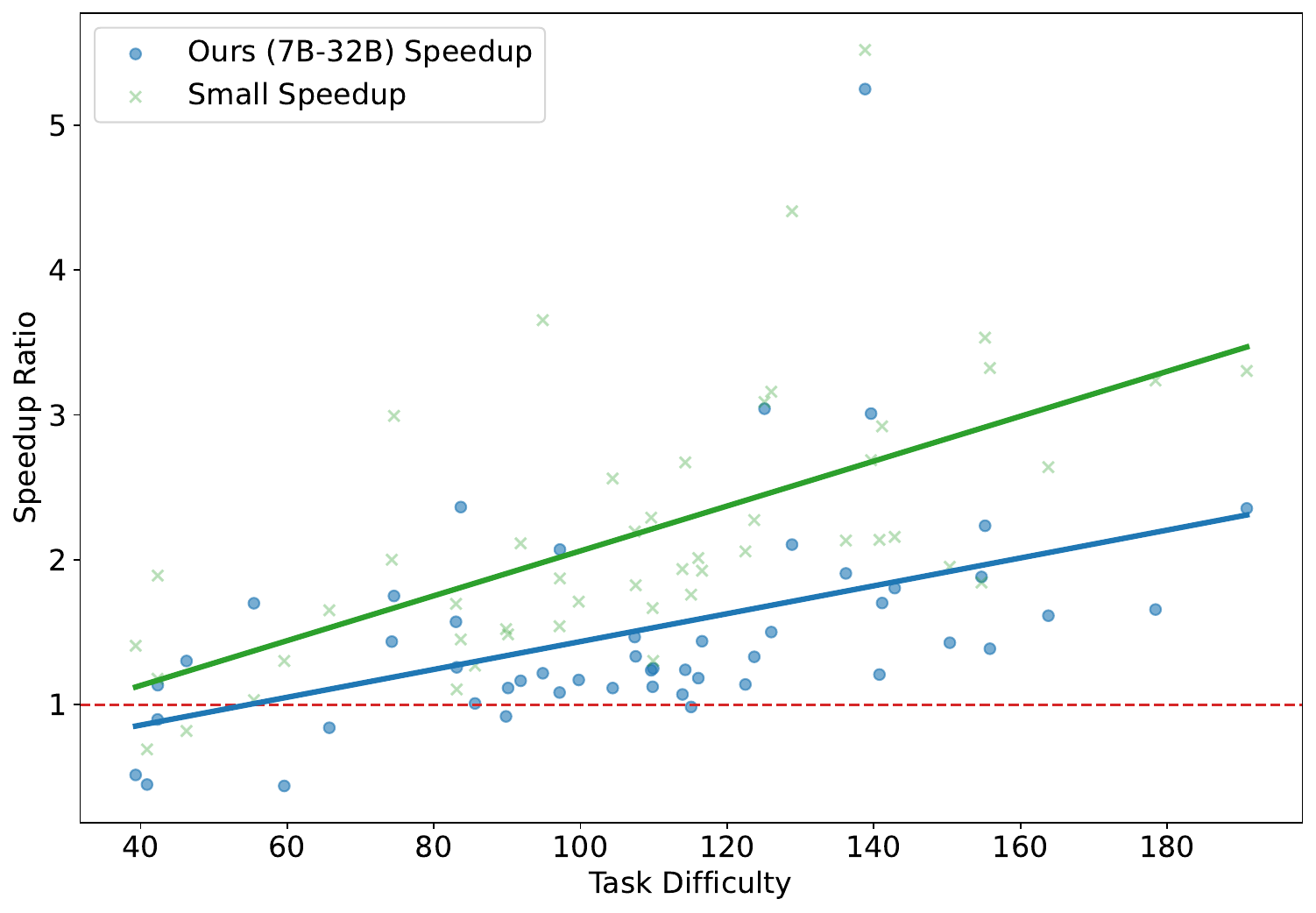}

\caption{
\textbf{Speedup ratio as a function of task difficulty.} Task difficulty is measured by the total latency of the Large (32B) baseline for each task (x-axis), and the y-axis reports the corresponding speedup ratio. Blue circles denote the speedup of Ours (7B--32B), while green crosses denote the speedup of the Small-only model (7B). Solid lines show linear trend fits. The red dashed line at $y=1$ indicates parity with the Large baseline.  The proposed AgentCollab maintains a clear positive trend under a selective large-model allocation strategy.
}
  
  \label{fig:difficulty_speedup}
\end{figure}

Figure~\ref{fig:difficulty_speedup} shows that the speedup of both AgentCollab and the small-only baseline increases with task difficulty. 
This trend suggests that harder tasks provide more room for reducing latency relative to the large-only setting. 
The small-only baseline exhibits a steeper increase, which is expected because it always executes the smaller model and therefore achieves greater latency reduction as the number of steps increases. 
By contrast, AgentCollab selectively invokes the large model when self-evaluation indicates insufficient progress. 
This design moderates the absolute speedup gain, but it preserves the corrective capability of the large model on challenging trajectory segments.

These results support the intended trade-off of AgentCollab. 
Rather than maximizing speed alone, the framework maintains substantial efficiency gains as task difficulty grows, while reserving stronger computation for cases in which purely small-model execution is more likely to fail.
\section{Effect of Model Size Pairing.}
To further examine how the choice of model sizes affects collaboration performance, we compare different small--large model pairings on WebSailor over BrowseComp\_zh, as shown in Table~\ref{tab:websailor_model_sizes}. Here, configurations such as 3B + 7B denote collaboration between two models of different sizes. For fair comparison, speedup is computed relative to the 7B baseline in the upper block and relative to the 32B baseline in the lower block.

As shown in Table~\ref{tab:websailor_model_sizes}, collaboration becomes more effective when the capability gap between the paired models is larger. In particular, 7B + 32B substantially improves over 7B alone and approaches the performance of 32B, while requiring fewer reasoning steps and achieving 1.50$\times$ speedup. By contrast, the gain from 3B + 7B is relatively limited, suggesting that the benefit of collaboration depends not only on model combination itself, but also on whether the stronger model can provide sufficient corrective capacity over the weaker one.
\section{Pseudo-code of AgentCollab}
\label{app:pseudocode}

Algorithm~\ref{alg:agentcollab} presents the full inference procedure of AgentCollab with static budget allocation strategy. 
The process consists of an initial planning stage with the large model, followed by a collaborative reasoning stage in which the small model handles routine progress and the large model is invoked only when self-evaluation indicates stagnation. 
During each escalation episode, the number of consecutive large-model steps is bounded by a predefined budget to avoid excessive computation. 
The procedure terminates once a final answer is produced from the accumulated context.
\section{Progress-check Prompt}
\label{app:progress_prompt}

For the progress check described in Section~\ref{sec:method}, we prepend the following instruction to the model's self-reflection step. This prompt requires the model to (i) summarize the newly observed actions and outcomes since the previous \texttt{THINK} step, (ii) assess whether these updates move the trajectory meaningfully closer to the final objective, and (iii) return a binary judgment in a strictly constrained output format.

\begin{algorithm*}[t]
\caption{AgentCollab with Self-Evaluation-Driven Escalation}
\label{alg:agentcollab}
\small
\KwIn{User query $Q$, Large model $M_L$, Small model $M_S$, Initial large-model budget $K_L$, Maximum consecutive large-model steps per escalation $B_L$}
\KwOut{Final answer $A$}

\tcp{Phase 0: Initialization}
$ctx \gets \text{init\_context}(Q)$\;
$mode \gets \text{LARGE}$\;
$large\_steps\_used \gets 0$\;

\tcp{Phase 1: Initial Planning with Large Model}
\While{$large\_steps\_used < K_L$}{
    $think\_out \gets M_L.\text{THINK}(ctx)$\;
    $ctx \gets ctx \cup \{think\_out\}$\;
    $prog \gets M_L.\text{PROGRESS\_CHECK}(ctx)$\;
    $large\_steps\_used \gets large\_steps\_used + 1$\;
    \If{$prog.value = \text{TRUE}$}{
        \If{$\text{is\_final\_answer}(ctx)$}{
            $A \gets \text{extract\_answer}(ctx)$\;
            \Return{$A$}\;
        }
    }
}
$mode \gets \text{SMALL}$\;

\tcp{Phase 2: Collaborative Reasoning with Self-Evaluation-Driven Escalation}
\While{\textbf{not} $\text{is\_final\_answer}(ctx)$}{
    \uIf{$mode = \text{SMALL}$}{
        $think\_out \gets M_S.\text{THINK\_AND\_TOOLS}(ctx)$\;
        $ctx \gets ctx \cup \{think\_out\}$\;
        $prog \gets M_S.\text{PROGRESS\_CHECK}(ctx)$\;
        
        \uIf{$prog.value = \text{TRUE}$}{
            \textbf{continue}\;
        }
        \Else{
            $mode \gets \text{LARGE}$\;
            $large\_steps\_used \gets 0$\;
            \textbf{continue}\;
        }
    }
    \ElseIf{$mode = \text{LARGE}$}{
        $think\_out \gets M_L.\text{THINK\_AND\_TOOLS}(ctx)$\;
        $ctx \gets ctx \cup \{think\_out\}$\;
        $prog \gets M_L.\text{PROGRESS\_CHECK}(ctx)$\;
        $large\_steps\_used \gets large\_steps\_used + 1$\;
        
        \uIf{$prog.value = \text{TRUE}$}{
            \If{$\text{is\_final\_answer}(ctx)$}{
                $A \gets \text{extract\_answer}(ctx)$\;
                \Return{$A$}\;
            }
            $mode \gets \text{SMALL}$\;
        }
        \ElseIf{$large\_steps\_used \ge B_L$}{
            $mode \gets \text{SMALL}$\;
        }
    }
}

\tcp{Phase 3: Finalization}
$A \gets \text{extract\_answer}(ctx)$\;
\Return{$A$}\;
\end{algorithm*}
\begin{figure*}[!htbp]
\centering
\begin{tcolorbox}[
    enhanced,
    width=\textwidth,
    colback=gray!8,
    colframe=black!80,
    boxrule=0.8pt,
    arc=2mm,
    title={Prompt for Progress Check},
    fonttitle=\bfseries,
    coltitle=white,
    colbacktitle=black!90
]

\begin{tcblisting}{
    listing only,
    colback=white,
    colframe=black!30,
    boxrule=0.4pt,
    arc=1mm
}
In this THINK step, I will first summarize what has changed since the previous THINK step,
then analyze whether these changes constitute significant progress toward the final objective.

Summary of new actions and results since the previous THINK step:
- 

Impact of these changes on progress toward the objective:
- 

Now I will make an explicit binary judgment about progress.

IMPORTANT: In the PROGRESS block below, I MUST strictly follow this structural format:
===PROGRESS===
<reason> one or two sentences explaining whether there is significant progress and why </reason>
<value>TRUE or FALSE only</value>
===END_PROGRESS===

and finish the generation immediately without any further content. Ignore the original requirements for the think tool.

Now I write the actual PROGRESS block following this structural format:
===PROGRESS===
<reason>
\end{tcblisting}

\end{tcolorbox}
\end{figure*}
% \section{Beyond the Paired Models}

% \begin{table}[!htbp]
% \centering
% \begin{tabular}{llll}
% \hline
% Size      & \textbf{Acc. (\%)} & \textbf{\#Steps} & \textbf{Speedup} \\ \hline
% 3B        & 9.7               & --               & --               \\
% 7B        & 14.2              & 10.38            & $2.03\times$     \\
% 32B       & 25.5              & 7.79             & $1.00\times$     \\
% 7B-32B    & 22.5              & 8.49             & $1.50\times$     \\
% 3B-7B-32B & --                & --               & --               \\ \hline
% \end{tabular}
% \caption{Preliminary extension of AgentCollab beyond paired models.}
% \label{tab:threemodel}
% \end{table}

% AgentCollab is not restrictfd to collaboration between only two models. Given $L$ models ordered by size, escalation can be performed progressively from smaller to larger models in a staged manner. In this setting, intermediate-capacity models can serve as transitional reasoning stages, allowing the system to allocate computation more gradually across different difficulty levels.

% \Cref{tab:threemodel} presents a preliminary example of this extension. To-Do: ANALYSIS .
\end{document}